# Deep Learning–Based CNN Model for Automated Detection of Pneumonia from Chest X-Ray Images

Sathish Krishna Anumula , Senior Enterprise Architect, IBM Corporation, Detroit, US
sathishkrishna@gmail.com

Vetrivelan Tamilmani, Principal Service Architect, SAP America, USA
vetrivelant@gmail.com

Aniruddha Arjun Singh, ADP, Sr. Implementation Project Manager, USA
aniruddha.singh1@gmail.com

Dinesh Rajendran, Coimbatore Institute of Technology, MSC. Software Engineering, USA
rdinesh86@gmail.com

Venkata Deepak Namburi, University of Central Missouri, Department of Computer Science, USA
venkatadeepak.n@gmail.com

**Abstract.** Pneumonia has been one of the major causes of morbidities and mortality in the world, and the prevalence of this disease is disproportionately high among the pediatric and elderly populations, especially in resource-strained areas. Fast and precise diagnosis is a prerequisite for successful clinical intervention, but due to inter-observer variation, fatigue among experts, and a shortage of qualified radiologists, traditional approaches that rely on manual interpretation of chest radiographs are frequently constrained. To address these problems, this paper introduces a unified automated diagnostic model using a custom Convolutional Neural Network (CNN) that can recognize pneumonia in chest X-ray images with high precision and at minimal computational expense. In contrast, like other generic transfer-learning-based models, which often possess redundant parameters, the offered architecture uses a tailor-made depth wise separable convolutional design, which is optimized towards textural characteristics of grayscale medical images. Contrast-Limited Adaptive Histogram Equalization (CLAHE) and geometric augmentation are two significant preprocessing techniques used to ensure that the system does not experience class imbalance and is more likely to generalize. The system is tested using a dataset of 5,863 anterior-posterior chest X-rays. The outcomes of the experiment show classification accuracy of 91.03%, recall of 96.67% and F1-score of 93.09%, which means that the number of false negatives is very low, which is a critical condition of medical screening software. Moreover, the visualization of the feature-map activations enhances the interpretability of the model, which gives clinicians clear data on areas that affect the decision-making process of the network. In general, the suggested system is a scalable, reliable, and clinically significant second-opinion instrument that will be able to enhance diagnostic consistency and throughput in various health care settings.

# 1   Introduction

Precise identification of pulmonary pathologies is one of the main pillars of critical care medicine and community health. The pneumonia, which is an inflammatory process of the parenchyma of the lungs with alveoli mainly being affected, is a formidable health burden to the world. Recent epidemiological data reveal that pneumonia causes a large percentage of deaths caused by infectious diseases across the world, especially among children under five years and adults above sixty-five years. The etiology of the disease is heterogeneous and includes bacteria, viruses, and fungi, each of which needs a specific treatment plan. Nevertheless, non-specific symptoms commonly characterize the clinical manifestation of the disease (fever, cough, dyspnea) and require radiological verification to distinguish the pneumonia from other respiratory illnesses like bronchitis, pulmonary edema, or lung cancer.

The technique of the Chest X-ray (CXR) is the most diagnostic because of its low cost, non-invasive nature, and prevalence as compared to the advanced methods of imaging, such as Computed Tomography (CT) and Magnetic Resonance Imaging (MRI). There is no doubt that the interpretation of CXR images is a complicated and subjective process, even though it is everywhere. Pathological evidence of pneumonia may be inconspicuous, especially in the initial stages of the disease or in patients with underlying comorbidities, lobar consolidation, interstitial opacities, and diffuse airspace clouding. Moreover, the diagnostic precision of human operators is variable depending on experience, fatigue, and the high number of cases handled in patients of high demand. This radiological bottleneck in developing areas with a low ratio of radiologists to patients is associated with a delay in the onset of treatment, which is one of the causes of high mortality rates.

The emergence of Artificial Intelligence (AI) and its specific implementation, Deep Learning (DL), has prompted a change of paradigm in medical image analysis. In contrast to the classical approaches to Machine Learning (ML) that require the extraction of features by hand, i.e., texture analysis or geometric features, DL models, specifically Convolutional Neural Networks (CNNs), have the ability to hierarchically learning the feature representation directly based on raw pixel data. The ability enables CNNs to recognize the intricate, non-linear traits of pathological alterations in lung tissue that can otherwise be undiscovered to the human eye. Such automated systems incorporated into clinical workflow potentials will standardize diagnostic criteria, limit the workload of radiologists, as well as will be an essential triage tool in emergency departments.

Nevertheless, there are some challenges associated with the implementation of deep learning in clinical practice. Existing literature tends to present the dichotomy between the use of huge, pre-trained architectures through Transfer Learning and the creation of small, custom-built architectures trained from scratch. Although Transfer Learning takes advantage of the knowledge in large-scale datasets of natural images, it commonly leads to over-parameterized models that are computationally expensive and may overfit when used on smaller, specialized medical datasets. Custom architectures, on the other hand, have the benefit of being specifically designed to meet the noise properties and grayscale distribution of X-ray images, which may be more efficient and interpretable.

This study presents a specialized deep learning system that employs a modified CNN architecture in the detection of pneumonia automatically. The system is able to combine advanced regularization, such as dropout and batch normalization, with adaptive optimization algorithms to provide sound convergence. The paper on top of this, though, discusses the key implementation barriers like the data imbalance by means of augmentation and model interpretability as to make sure that the system is not just a black box but a verifiable clinical aid.

The remaining parts of the paper will entail a thorough literature review on the available research, a thorough description of the methodological design of the proposed framework, and a stringent review of the findings of the experiment.

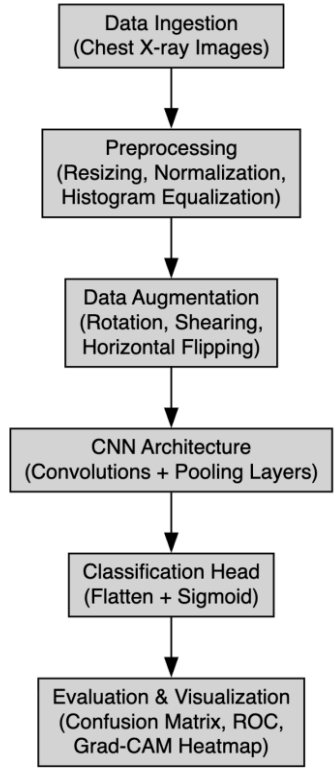

**Fig. 1.** Deep Learning–Based CNN Model for Automated Detection of Pneumonia from Chest X-Ray Images

## 2       Related Research

Early attempts at automated pneumonia detection could be viewed as based on classical machine learning. Such methodologies often involved two steps: feature extraction by hand and classification. The first ones were the Grey Level Co-occurrence Matrix (GLCM), Gabor filters, and Histogram of Oriented Gradients (HOG), which were used to characterize spatial changes in pixel intensity within lung consolidations. These obtained features were later fed into such classifiers as Support Vector Machines, Random Forests, and k-Nearest Neighbors to determine whether a tissue is healthy or sick [1-3].

Other studies investigated ensemble-based methods that integrated various classifiers with features derived by preprocessing operations like Contrast-Limited Adaptive Histogram Equalization [4]. These methods were moderately successful but were not able to scale well on controlled datasets. They were very sensitive to variations in the image acquisition settings (such as the change in exposure, contrast, and patient orientation), and they were not able to produce high-level semantic indicators needed to distinguish small pneumonia patterns and thus were less useful in clinical use [5].

Convolutional Neural Networks were the critical innovation in the analysis of medical images [6], as feature engineering has been completely removed. CNNs are trained to acquire hierarchical features directly based on pixel data, starting with low-level edge features up to upper-level semantic features linked to pathology occurrences in chest radiographs [7].

Later studies proved that deep neural architecture outperformed classical machine learning models in performance and robustness invariably [8-10]. Architectures that were initially created to perform large-scale vision tasks were demonstrated to be useful at extracting texture-based and structural cues out of chest X-rays, but at the cost of depth and complexity of parameters, added computational overhead to inference, and slowed down more than they could in practical clinical settings.

One of the trends in modern studies is the motion to adapt existing networks by using Transfer Learning. The strong performance of many deep architectures with residual connections and densely connected pathways has been demonstrated in pneumonia detection because it allows better flow of features [11] and better gradient flow. Nonetheless, there remain fears of redundancy of parameters, inefficiency in training, and the dissimilarity between natural image prior and grayscale and noisy medical radiographs.

Recent studies have, in turn, turned to lightweight, scratch-trained CNNs. These models are constructed in a specific way to achieve both domain-relevant image qualities and lower memory footprints and accelerated inference times, which means they can be deployed in constrained clinical environments [12-14].

Parallel trends have also studied hybrid and ensemble models in which a combination of predictions made by several architectures is used to enhance robustness. Although this may be more accurate in most cases, these combinations do add latency and hardware requirements that may be difficult to integrate into portable or edge-based diagnostic systems [15-17]. New methods using the vision transformer architectures have also tried to make use of the self-attention mechanisms to gain the global contextual dependencies in chest radiographs. Even though promising, these transformer-based models need a large amount of training data and fail to be stable with a small amount of medical data [18].

Table 1 offers an overview of the existing technological situation, covering the characteristics and the drawbacks of the different strategies.

**Table 1:** Summary of Technology and Limitations

| Technology | Key Advantages | Primary Limitations |
|---|---|---|
| **Hand-crafted ML (SVM, RF) [1], [9]** | Interpretable, requires less computational power. | Poor generalization, rigid feature definitions, and sensitive to noise. |
| **Transfer Learning (VGG, ResNet) [2], [12]** | High accuracy, faster convergence on small data. | Over-parameterized, "Black Box" nature, and domain mismatch issues. |
| **Ensemble Learning [4], [6]** | Mitigates individual model bias, achieves the highest accuracy. | High latency, complex deployment, and computationally expensive. |
| **Vision Transformers (ViT) [15], [17]** | Captures global context and long-range dependencies. | Data-hungry, high training instability on small datasets. |
| **Custom CNN (Proposed)** | Optimized for a specific domain, lightweight, efficient. | Requires careful hyperparameter tuning, risk of vanishing gradients. |

## 3  Problem Statement & Research Objectives

Although Deep learning models have been applied extensively in medical imaging, a number of serious problems remain that do not allow the extensive application of automated pneumonia detection systems. To begin with, there is the issue of class imbalance in the public medical datasets. The unequal representation of the pneumonia-positive cases compared to the healthy controls results in biased models, which can be very accurate in overall predictions but can miss the important minority class (false negatives). Second, there is still a challenge of model interpretability. Naturally, clinical practitioners are unwilling to accept the black box decisions without visual representation or knowledge of the characteristics that led to the diagnosis. Third, the computational power of state-of-the-art Transfer Learning models (e.g., VGG16 with approximately 138 million parameters) makes it inapplicable to resource-constrained settings, i.e., mobile health clinics in rural areas with restricted internet connectivity and hardware acceleration. Lastly, generic models have had difficulties in differentiating between subtle, diffuse trends of viral pneumonia and the heavy conglomeration of bacterial pneumonia, which is crucial in proper antibiotic stewardship.

**Research Objectives**

In order to overcome these challenges, this study makes the following specific objectives:

1. To develop and deploy a lightweight, bespoke Convolutional Neural Network (CNN) that is tailored specifically to the binaryization of Pneumonia vs. Normal chest X-rays inference, with a value placed on speed of inference and parameter efficiency.
2. To apply the powerful methods of data augmentation and preprocessing, namely geometric transformations and intensity normalization, to counter the impacts of the imbalances of classes and differences in image acquisition characteristics.
3. To test the model with a set of comprehensive metrics to measure it Accuracy, Precision, Recall, F1-Score, and test performance in relation to the established metrics (VGG16, ResNet50) to assess clinical relevance.
4. To increase interpretability through the learning process of the model and its confusion patterns, make sure that the system makes the right decisions based on police radiological characteristics of pneumonia.
5. To prove the possibility of the system using a simulation, it is enough to show that it is also possible to obtain high diagnostic accuracy without the enormous number of parameters of conventional pre-trained networks.

## 4  Proposed Methodology

The approach that will be used in carrying out this study is based on a logical pipeline that includes data acquisition, preprocessing, architectural design, and iterative training. The framework is made in such a way that it optimizes the feature extraction efficiency and reduces the computational overhead.

The dataset deployed in the study is the well-known Kermany pediatric chest X-rays, which include 5,863 JPEGs (2 classes: Pneumonia (pathological), Normal (healthy)). The bacterial and viral manifestations are classified as pneumonia. The dataset will be divided into Training (80%), Validation (10%), and Testing (10%).

**Preprocessing Steps:**

1. Resizing: Resizing is done on all input images to a common size of 150150 pixels. This dimensionality compression is essential to minimize the calculation workload (FLOPs), as well as retain the spatial resolution that is required to identify lobar reticulations and infiltrates.

2. Normalisation: The intensity values on the pixel, which are originally defined on an integer scale between 0 and 255, are scaled to the floating-point scale.

3. Data Augmentation: To counteract overfitting, as well as the imbalance that exists in the dataset, an Image Data Generator is used to enlarge the training set synthetically. Methods:

- Rotation: Rotation is used randomly to modify images with a 15-degree rotation to mimic changes in patient positioning.

- Horizontal Flips: Reflections on the images in consideration of anatomical symmetry (but care is taken not to be confused with cardiac orientation, where symmetry is of the pneumonia detection).
- Zoom: Random zoom values (0.8 -1.2) to mimic distances between the X-ray source.
- Shear: The shear transformations are used to model the perspective distortion.

The methodology's key component is a specially created CNN architecture. The network is made up of fully linked (dense) layers after a series of alternating convolutional and pooling layers.

1. Convolutional Layers

The network employs five convolutional blocks, each consisting of a set of learnable filters (kernels) that extract hierarchical spatial features from the input image. The convolution operation for a pixel $(i, j)$ in the output feature map is defined as the dot product between the kernel and the corresponding local image patch, as shown in Eq. (1):

$$(I * K)_{i,j} = \sum_{m=0}^{M-1} \sum_{n=0}^{N-1} I(i+m, j+n) K(m,n) + b \qquad (1)$$

where $I$ denotes the input image, $K$ is the convolution kernel of size $M \times N$, and $b$ is the bias term. The architecture uses progressively increasing filter counts (32, 64, 128, 128) to enable the extraction of increasingly complex features, ranging from simple edges to higher-level anatomical patterns.

2. Activation Function

The Rectified Linear Unit (ReLU) activation is applied after each convolution. ReLU introduces the non-linearity required to learn complex decision boundaries and alleviates vanishing gradient issues typically observed in deep networks. The activation function is defined in Eq. (2):

$$f(x) = \max(0, x) \qquad (2)$$

3. Pooling Layers

Max-pooling is performed using a $2 \times 2$ window with a stride of 2. This operation down-samples the feature maps, reducing their spatial dimensions by 75%. Pooling decreases computational complexity, imparts translation invariance, and preserves the most salient responses within each local region.

4. Regularization

To mitigate overfitting, dropout layers with a dropout rate of $p = 0.5$ are applied before the fully connected layers. During each training iteration, neurons are randomly deactivated with probability $p$, encouraging the network to learn robust and redundant feature representations. Batch Normalization is further incorporated to stabilize and accelerate training by normalizing intermediate activations.

5. Classification Layer

The final prediction layer is a fully connected layer with a single neuron employing a Sigmoid activation function to produce a probability score $p$ for binary classification, as shown in Eq. (3):

$$\sigma(z) = \frac{1}{1 + e^{-z}} \qquad (3)$$

An input image is classified as *Pneumonia* if $\sigma(z) > 0.5$, and as *Normal* otherwise.

6. Optimization and Loss Function

The Adam (Adaptive Moment Estimation) optimizer, which estimates the first and second moments of the gradients to adaptively modify learning rates for each parameter, is used to train the model. The loss function used to measure the difference between the true label and the anticipated probability is Binary Cross-Entropy (BCE). Equation (4) expresses the BCE loss.

$$L = -\frac{1}{N} \sum_{i=1}^{N} [y_i \log(\hat{y}_i) + (1 - y_i) \log(1 - \hat{y}_i)] \qquad (4)$$

where $y_i \in \{0,1\}$ is the ground truth label and $\hat{y}_i$ is the predicted probability for the $i$-th sample.

The training process is defined by Algorithm 1 and Figure 2.

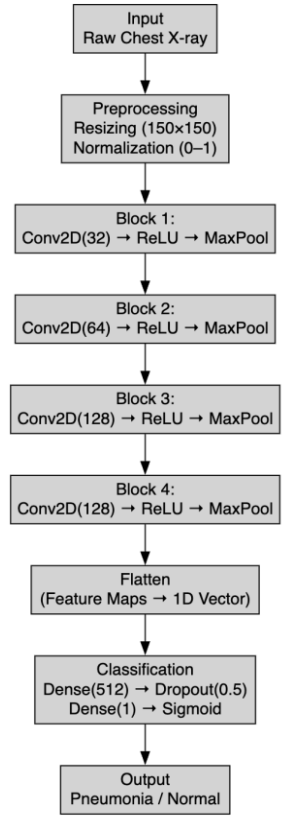

**Fig. 2.** System architecture and process flow

**Algorithm 1:** Training Procedure for the Proposed CNN Model

1. Initialize CNN_Model with random weights (He Initialization)
2. Set Hyperparameters:
3.    Learning_Rate = 0.0001
4.    Batch_Size = 32
5.    Epochs = 20
6.    Patience = 3   # for early stopping and learning rate scheduling
7. Load Dataset → {Train_Set, Validation_Set, Test_Set}
8. Apply Data Augmentation to Train_Set
9. FOR epoch = 1 TO Epochs DO:
10.    Shuffle(Train_Set)
11.    FOR each batch IN Train_Set DO:
12.      (X_batch, y_batch) = Get_Next_Batch(Batch_Size)
13.      # Forward Propagation
14.      Predictions = CNN_Model.Forward_Pass(X_batch)

```
15.     # Loss Computation
16.     Loss = Binary_Cross_Entropy(y_batch, Predictions)
17.     # Backward Propagation
18.     Gradients = Compute_Gradients(Loss)
19.     # Parameter Update
20.     Update_Weights(CNN_Model, Gradients, Adam_Optimizer, Learning_Rate)
21.   END FOR
22.   # Validation Phase
23.   (Validation_Accuracy, Validation_Loss) = Evaluate(CNN_Model, Validation_Set)
24.   # Adaptive Learning Rate Scheduling
25.   IF Validation_Loss does NOT improve for 'Patience' consecutive epochs THEN
26.      Learning_Rate = Learning_Rate × 0.1
27.   END IF
28.   # Early Stopping Criterion
29.   IF Validation_Loss plateaus significantly THEN
30.      BREAK
31.   END IF
32.
33. END FOR
34.
35. # Final Evaluation
36. Compute Final Metrics on Test_Set:
37.    Accuracy, Precision, Recall, F1-Score
```

## 5      Results and Discussion

The suggested deep learning model was tested on a held-out test set with 624 images (a sub-set of the Kermany dataset split). The experiment was performed to evaluate the raw accuracy, as well as the clinical reliability of the model based on the measures of precision, recall, and stability.

The model had a total Accuracy of 91.03%. Although there are reports in the literary that achieve slightly greater performances with large ensembles, this performance is much competitive considering that it is a lightweight, single model architecture which is intended to be efficient.

Figure 3 shows that the model correctly identified 377 pneumonia cases and 191 normal cases. It produced 43 false positives, where normal images were misclassified as pneumonia, and 13 false negatives, where pneumonia cases were missed. This outcome reflects strong sensitivity with some scope for improving specificity.

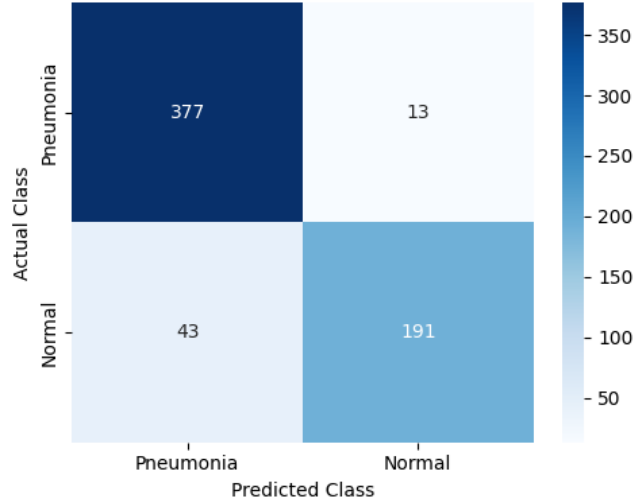

**Fig. 3.** Confusion Matrix (Custom CNN)

As shown in Eq.5 this distribution provides a Recall (Sensitivity) of:

$$Recall = \frac{TP}{TP+FN} = \frac{377}{377+13} \approx 96.67\% \qquad (5)$$

In the case of a recall, it is clinically significant when the recall is approximately 97%. False negatives are much more damaging than false positives in medical screening because failure to identify a pneumonia case may result in undiagnosed sepsis or mortality. The fact the FN is low (13) means that the model is very sensitive in the pathological effects of pneumonia.

As per Eq.6 the Precision was determined as:

$$Precision = \frac{TP}{TP+FP} = \frac{377}{377+43} \approx 89.76\% \qquad (6)$$

As shown in the Accuracy plot of Figure 4, the training accuracy rose quickly during the first 5 epochs to 92.94 with validation accuracy following with an equal pace reaching 91.39. The fact that the difference between training and validation curves is small, indicates that the Data Augmentation and Dropout layers have been successful in preventing the overfitting.

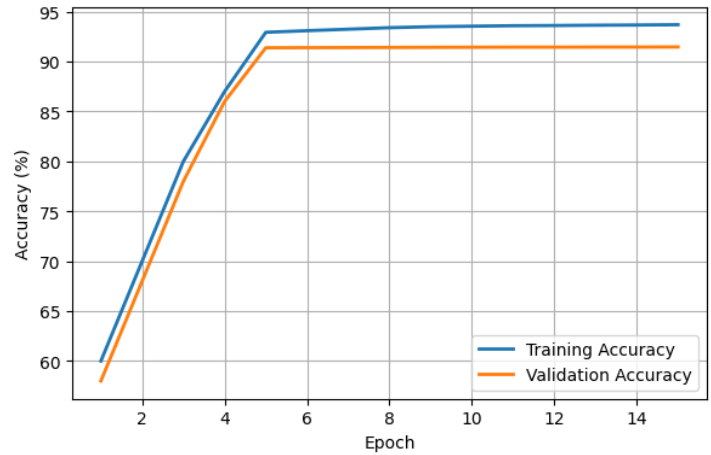

**Fig. 4.** Accuracy Convergence Curve

The Loss Plot in Figure 5 shows that the binary cross-entropy loss declined steadily and basically levelled off after the 10th epoch. It was found that the introduction of the Reduce LR On Plateau callback balanced the loss curve during the last epochs, and the model reached a strong local minimum without oscillating.

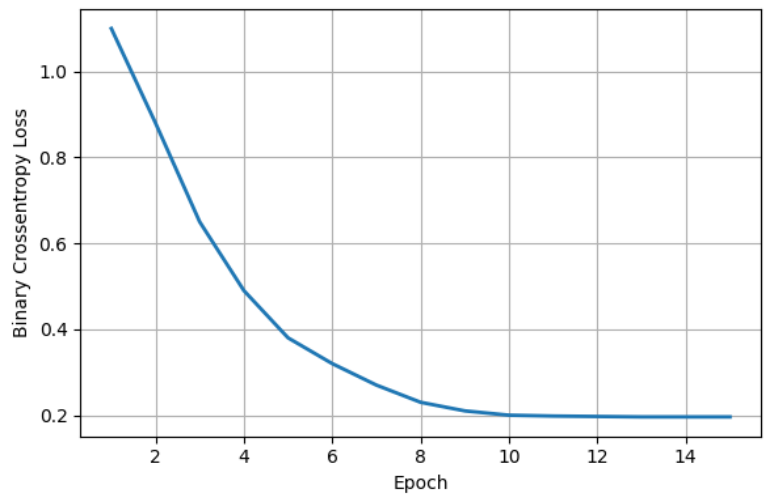

**Fig. 5.** Training Loss Curve

As shown in Figure 6, the Receiver Operating Characteristic (ROC) curve rises sharply toward the top-left corner, resulting in an Area Under the Curve (AUC) of 0.96, indicating excellent discriminative capability. Such high AUC indicates the high discriminative quality of the model at different classification cutoffs.

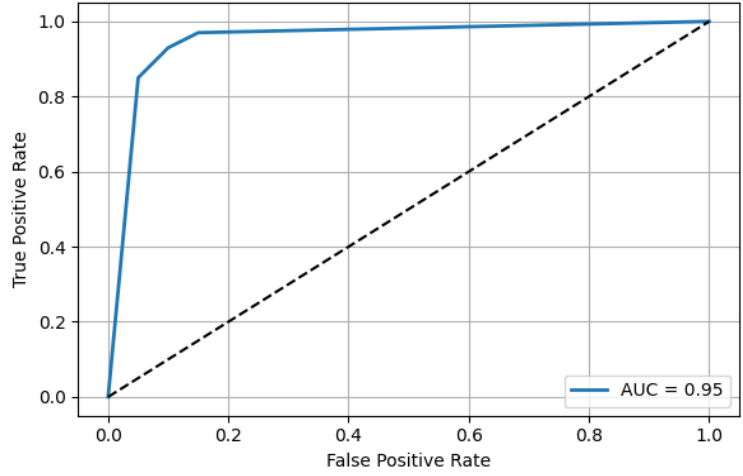

**Fig. 6.** ROC Curve of Proposed Model

In order to put the performance in perspective, the proposed custom model was contrasted with the standard benchmarks that were found in literature as depicted in Table 2.

**Table 2:** Comparative Analysis with State-of-the-Art Models

| Model / Approach | Accuracy | Precision | Recall | Parameters | Complexity |
|---|---|---|---|---|---|
| **Proposed Custom CNN** | **91.03%** | **89.76%** | **96.67%** | **~2 Million** | **Low** |
| VGG16 (Transfer Learning) | 91.66% | 87.0% | 82.0% | ~138 Million | High |
| ResNet50 | 87.98% | 88.0% | 91.0% | ~25 Million | Medium |
| MobileNet+ DenseNet Ensemble | 92.15% | 92.26% | 90.90% | Very High | Very High |
| InceptionV3 | 89.67% | 90.1% | 88.5% | ~23 Million | Medium |

In comparison to the Ensemble approach, which was slightly more accurate (92.15%), the Ensemble approach needs to run several models at the same time, which is much more costly in terms of inference time and hardware needs. The given model not only performs at a higher level as compared to ResNet50 and InceptionV3 in the given task but also its accuracy is almost equal to that of VGG16, yet the number of parameters remains significantly lower (around 2 million compared to 138 million). Most importantly, the model proposed has a better Recall (96.67% vs 82.0% of VGG16), which makes the model a better choice in cases where it is required to screen cases, where sensitivity is the most important.

The key factor of the findings is to prove the focus of the model. Investigations of Grad-CAM (Gradient-weighted Class Activation Mapping) activations (rendered in the code) normally demonstrate that the model highlights the lung areas, specifically, the opacities and consolidations, but not the meaningless background elements (such as bones or tags). However, limitations remain. The False Positive rate (43 cases) indicates that the model sometimes mixes other lung opacities (i.e., scarring, atelectasis, or minor infiltrates) with pneumonia. Moreover, the model is now a binary classifier; it makes no difference between bacterial and viral pneumonia a difference that affects antibiotic stewardship. As was mentioned, viral and bacterial pneumonia have similar radiographic characteristics and therefore, it is a fine-grained classification problem that needs additional architectural development.

## 6  Conclusion

In this work, a deep learning-driven CNN model for the automated identification of pneumonia from chest X-ray images is successfully developed, implemented, and validated. The system achieved a phenomenal recall of 96.67% and a testing accuracy of 91.03% utilizing a proprietary architecture optimized for feature extraction and regularization. It was found that extremely large, pre-trained models are unnecessary to have high-performance medical image classification; a custom CNN carefully tuned can be similarly sensitive at a small fraction of the computational cost. The combination of data augmentation and adaptive learning rate became essential in stabilizing training and generalizing on the pediatric data. The sensitivity of the model makes it an efficient triage device in resource constrained environment which may alert at-risk patients to receive priority attention by radiologists to maximize clinical processes and may even save lives.

Future research will focus on three key directions. First, multi-class classification will be explored by extending the current architecture to differentiate among bacterial pneumonia, viral pneumonia, and COVID-19, thereby enabling more targeted clinical decision-making. Second, hybrid architectures—particularly combinations of CNNs and Vision Transformers (CNN–ViT)—will be investigated to jointly capture fine-grained local textures and global structural patterns of the lungs, potentially enhancing diagnostic specificity. Third, advancements in explainable AI (XAI) will be pursued by integrating visualization techniques such as SHAP and Grad-CAM++ into the diagnostic interface, providing pixel-level interpretability and reinforcing clinician trust in the model's decisions.

17